# AI and Pathology: Steering Treatment and Predicting Outcomes

Rajarsi Gupta, Jakub Kaczmarzyk, Soma Kobayashi,
Tahsin Kurc, Joel Saltz

Department of Biomedical Informatics
Stony Brook University

## 1  Introduction

The human body is constructed from over $10^{13}$ cells organized in complex patterns of hierarchical organization. Each cell carries out its own metabolism. During various portions of the human lifespan, different types of cells replicate, differentiate, migrate, and sometimes die. The combination of AI, high-end computing capabilities, and improvements in sensor and molecular characterization methods are making it possible to analyze humans and other animal and plant life forms in a quantitative granular, multi-scale, cell-based fashion.

We focus on a particular class of targeted human tissue analysis - histopathology - aimed at quantitative characterization of disease state, patient outcome prediction and treatment steering. During the past 150 years, much of anatomic pathology consisted of characterization and diagnostic recognition of visually identifiable morphological patterns seen in tissue. Over the past few decades, molecular characterizations have become increasingly important in anatomic Pathology. The advent of AI and ubiquitous high-end computing are enabling a new transition where quantitatively linked morphological and molecular tissue analyses can be carried out at a cellular and subcellular level of resolution. Along with quantitative assessments of predictive value and reproducibility of traditional morphological patterns employed in anatomic pathology, AI algorithms are enabling exploration and discovery of novel diagnostic biomarkers grounded in prognostically predictive spatial and molecular patterns.

## 2  Digital Histopathology: A Brief Background

Histopathology is best defined by understanding its two root words: *histology* – the microscopic study of tissue structures – and *pathology* – the study of diseases. Histopathology thus involves the microscopic analysis of patient tissue specimens and changes in tissue structure to understand disease causes and mechanisms. Pathologists analyze clinical tissue specimens under powerful microscopes to evaluate changes in tissue structure. Based on their evaluation, they diagnose, subtype, determine prognosis, and help guide treatment decisions of patients. Acquisition of tissue samples is the first step of this process and is largely performed from organ sites with active disease. Some examples are tissue taken directly from cancer or from intestinal regions affected with inflammatory bowel disease. Several additional steps are taken before collected tissue is ready for microscopic evaluation. The first involves fixation, either in formalin or by freezing, to preserve the tissue and structural architecture. Fixed tissues are sliced into thin sections of several microns in thickness onto glass slides. Finally, the tissue sections are treated by a process called staining that allows structures to be more emphasized and visible to the human eye. Hematoxylin and Eosin (H&E) staining, for example, reveals features such as organization, texture, and morphology of cell nuclei and the multicellular tissue structures they form together. Cancer can be diagnosed by the replacement of expected cell types with aberrant-appearing cancer cells and the disruption of typical organ-specific multicellular organization. For certain diagnoses, the spatial distribution of molecular markers is crucial. Another type of staining, called Immunohistochemistry (IHC), applies antigen-specific antibodies that target specific markers within tissue specimens. This identifies the presence or absence of microscopic protein targets in the tissue. Both structural and molecular information are employed in determining cancer subtype and in guiding treatment.

A pathologist's examination of a tissue slide under a microscope is a *qualitative* process. The pathologist will use their experience and pathology knowledge to classify tissue using complex classification criteria developed over the years by both the Pathology and the broader medical community [1]. The classifications evolve over time in response to a combination of observational clinical studies and clinical trials. The interpretive process is subjective [2–4]; different Pathologists who examine the same specimen may not classify the tissue in the same way.

Digital histopathology facilitates *quantitative* analyses of tissue specimens. Tissue acquisition and preparation processes (tissue fixation, staining protocols) are the same as in traditional pathology. A glass tissue slide with H&E or IHC staining is placed in a digital microscopy instrument with state-of-the-art imaging sensors. These instruments capture an image of the tissue specimen like a digital camera, but at much higher resolution to generate whole slide images (WSI) that can be more than 100,000x100,000 pixels in size. WSIs can be generated at different levels of resolution; a pixel will commonly represent a 0.25 to 1 micron tissue region.

Digital microscopy instruments have improved remarkably over the past 20 years. Scanning a tissue specimen at 0.5 micron per pixel resolution used to require several hours. Modern instruments can output WSIs at 0.25 micron per pixel resolution in approximately one minute. They have advanced focusing mechanisms and automated trays which can hold hundreds of glass slides, enabling high throughput generation of WSIs in a pathology lab or a tissue bank. These systems have increasingly become ubiquitous in research settings and are now being widely adopted for clinical use [5,6]. As the digital microscopy instruments get faster, higher resolution and more cost effective, we expect tissue banks and pathology labs at research and clinical institutions will be able to scan hundreds of thousands of glass slides per year in the not-too-distant future. Indeed, the BIGPICTURE project of the EU Innovative Medicines Initiative will collect millions of digital slides for development of AI methods as well as AI applications in pathology research[7].

GLOBOCAN 2020[8] estimates of cancer incidence and mortality produced by the International Agency for Research on Cancer. Worldwide, an estimated 19.3 million new cancer cases (18.1 million excluding nonmelanoma skin cancer). The number of slides examined per patient depends on the nature of the cancer, local treatment practices and available resources. Worldwide, a conservative estimate of five to ten pathology tissue slides per new cancer patient yields a rough worldwide estimate of 100-200 million tissue specimens created per year. Many additional slides are generated in diagnostic workups that do not lead to a cancer diagnosis and in assessment of cancer recurrences and metastases.

## 3    Applications of AI in Digital Histopathology

Gigapixel resolution WSIs offer a rich landscape of data that can be leveraged by computation to improve diagnostic reproducibility, develop high resolution methods for predicting disease outcome and steering treatment, and reveal new insights to further advance current human understanding of disease. In this work, we look at the applications of AI in three important tasks that support these goals. The first task is the extraction of quantitative features from WSIs. Quantitative imaging features can enable novel insights into pathophysiological mechanisms at disease onset, progression, and response to treatment. This task generally involves the process of detecting, segmenting, and classifying objects (e.g., nuclei and cells), structures (e.g., stroma) and regions (e.g., tumor regions) within a WSI. Figure 1 shows an example of a breast tissue specimen being segmented into regions with invasive breast cancer and normal tissue. Figure 2 shows results of a model that segments invasive cancer and regions with infiltrating immune cells. The second task is the classification of WSIs. Classifications in Pathology are complex and nuanced with hierarchies consisting of tens of thousands of sub-categories. The third and overarching task is to predict patient outcome and to assess the impact of different treatments on a patient's disease course. This task differs from the first two as it involves prediction of intrinsically uncertain future events. While predictions can be carried out with digital pathology information alone, in a broad sense, the best predictions are likely to require integrated analysis of microscopic tissue images, Radiology images, clinical and molecular information.

### 3.1 Computer Science Challenges

Viewed from a computer science perspective, computer vision tasks in digital pathology include: 1) detection and segmentation of objects in high-resolution image data, 2) assignment of a class label to each image in an image collection and 3) predictions from a combination of image and other data sources. The work on panoptic segmentation [9] abstracts key requirements associated with many digital pathology segmentation-related tasks. In some cases, object detection and quantification rather than segmentation are all that an application requires. The computer vision community has developed a variety of algorithms to address such problems; see for instance [10]. Nevertheless, segmentation is a challenging task in digital pathology, because the tissue landscape is highly heterogeneous and complex. There are hundreds of thousands of cells and nuclei and thousands of micro-anatomic structures and regions in a typical tissue specimen. The texture, shape, and size properties of cells, nuclei and structures can vary significantly across tissue specimens (depending on the cancer type/subtype and stage) and even within the same tissue. The task of assigning a class label to images in a collection is extremely well recognized in the computer vision community; there has been transformative progress in this area over the past decade. Digital pathology image classification problems are challenging as they typically involve gigapixel resolution images and frequently include classifications with substantial nuance. Creation of fine-grained training data is difficult, time consuming, expensive, and requires expert clinician insight. A major challenge is thus to train classification models using sparse, weakly annotated training data.

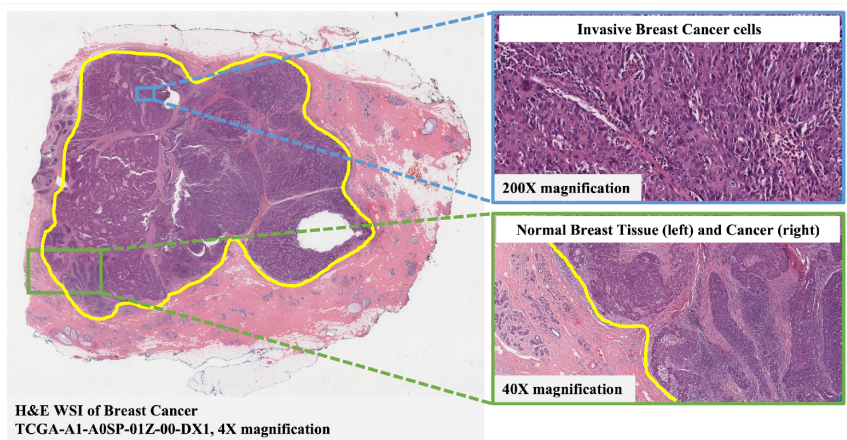

Figure 1. H&E WSI of breast cancer tissue sample from the Cancer Genome Atlas (TCGA). Insets show invasive breast cancer cells shown at 200X and a representative area showing the invasive margin between cancer and normal tissues delineated by the solid yellow line.

Prediction-related tasks include 1) prediction of which patients will develop a given disease, 2) prediction of how quickly and in what ways a given disease will progress and 3) prediction of how given patients will respond to a particular treatment. Predictions can be made using Pathology images alone, yet will likely need to progress to integrate analysis of temporal sequences of combined clinical observations, clinical notes, Radiology imaging, Pathology images, molecular data and treatment data. There are many different ways prediction problems can be formulated and solved with myriad computational and mathematical challenges associated with the coupled analyses of highly inter-related but structurally disparate imaging, genomic and clinical datasets.

### 3.2 Tools and Methodology

A wide spectrum of deep learning methods, such as convolutional neural networks (CNNs), multi instance learning (MIL), attention mechanisms, recurrent neural networks (RNNs), and transformers, have been developed by the data analytics community [11,12]. Digital pathology has adapted many of them. CNNs, for example, are routinely employed in the WSI segmentation and classification tasks described above. Multi-resolution analysis and MIL methods are frequently employed to carry out WSI classification. Transformers are being leveraged to generate image representations that can be incorporated into downstream classification and segmentation tasks. Development and application of AI methods for digital pathology image analysis is a highly active area of research. In the next sections, we provide examples of how AI has been applied in the three tasks described above. We refer the reader to excellent survey papers for a more comprehensive coverage (e.g., [13–15]).

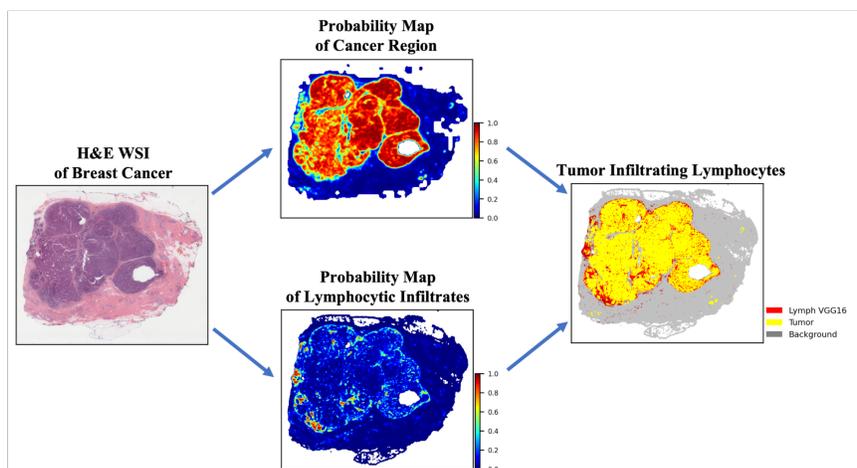

Figure 2. Segmentation and classification of breast cancer tissue and TILs. Breast cancer tissue segmentation is indicated by the yellow color, where superimposed red color shows the presence and spatial distribution of TILs at 50 um resolution. H&E WSI from TCGA Breast Cancer (BRCA) collection, specimen TCGA-A1-A0SP-01Z-00-DX1.

### 3.3 Detection, Segmentation and Characterization of Microanatomic Structures

Tissue is made up of micro-anatomic structures such as cells, ducts, stromal regions, and glands. The nucleus is a core micro-anatomic structure that harbors cellular DNA and plays a central role in disease diagnosis and prognosis. A telltale sign of cancer, for example, is abnormal nuclei. Cancer causes visible changes in the shape, size and texture of the nucleus (e.g., cancerous nuclei are generally larger and more irregularly shaped). Quantitative features computed from nuclei could provide novel biomarkers to study the pathophysiology of cancer and better understand its mechanisms. Cooper et al, for instance, used the shape and texture features of nuclei to create morphology-driven clusters of glioblastoma cases [14]. They observed correlations between the clusters and cancer subtypes and genomic signatures, indicating the diagnostic and prognostic power of quantitative nuclear features. In another study, Yu et al. computed thousands of features from segmented nuclei and cells from lung cancer cases [15]. A downstream machine learning analysis using the features showed correlations with patient outcome.

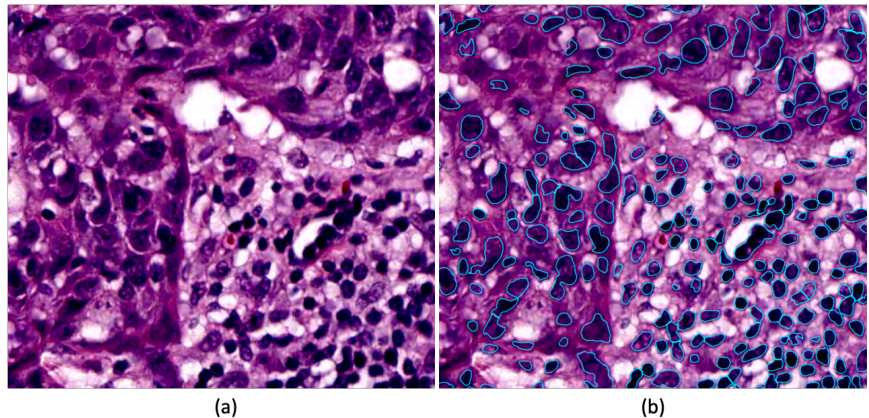

Figure 3. An example of nucleus segmentation: (a) An image patch with tissue and nuclei. (b) Nuclei segmented in the patch are shown in blue polygons.

The process of extracting quantitative nuclear features from tissue images requires the detection, segmentation and classification of nuclei. This can be a challenging task because of the volume and morphological variability of nuclei in tissue images. For example, lymphocyte nuclei are dark, round and small with 8 micron diameter on average. The nuclei of cancer cells, on the other hand, are generally larger than 10 microns in diameter and irregular in shape. Moreover, nuclei may touch or overlap each other without easily distinguishable boundaries. Figure 3 shows an example of nucleus segmentation and classification in a WSI. Kumar et al. developed a nucleus segmentation pipeline with a CNN architecture consisting of multiple convolution and fully connected layers [16]. They labeled pixels in the training dataset with one of three classes; inside a nucleus, outside a nucleus, and boundary pixel. They introduced the boundary pixel label to train a model that can separate touching nuclei. Their segmentation model was effectively a three-class pixel-level classifier. They determined the number of convolution and fully connected layers empirically by testing different configurations. After input images were processed by the CNN model to label nuclei and their contours, a region growing method was applied to the output to finalize the boundaries of nuclei. In a more recent work, the top scoring participants of a digital pathology challenge proposed a multi-step deep learning pipeline for nucleus segmentation [17]. The pipeline trained two CNNs to detect blobs of nuclei and their boundaries to generate blob and border masks. The masks were then used in the next step of the pipeline to separate nucleus boundaries by dilating the border mask and subtracting it from the blob mask. In the final stage, a watershed algorithm was used to identify individual nuclei.

In many works, classification of nuclei into classes (such as malignant epithelium, lymphocyte, endothelial) is handled as a post-processing step after nuclei are segmented and a set of features are computed. Graham et al. proposed a novel CNN architecture, called HoVer-Net, to both segment and classify nuclei in a single method [18]. The proposed architecture includes a feature extraction module with layers of residual units to extract deep features from input images. The features are then input to three CNN branches for segmenting individual nuclei and classifying them. These branches predict (1) if a pixel is part of a nucleus or not, (2) the horizontal and vertical distances of identified nucleus pixels to their centers of mass (for the purpose of separating touching nuclei), and (3) the type of nucleus for each nucleus pixel. The branches consisted of upsampling layers, which restore image resolution reduced in the feature extraction module, and densely connected layers for predicting nucleus pixels and boundaries. In an experimental comparison to other nucleus segmentation methods, HoVer-Net achieved state-of-the-art results in multiple accuracy metrics.

Deep learning methods require large amounts of high-quality training data to train accurate and generalizable models. This has been an ongoing challenge in nucleus segmentation work. It is time consuming to manually segment individual nuclei, and this process often requires input from expert pathologists to ensure nuclei and their boundaries are correctly identified. A number of projects have implemented crowd-sourcing and semi-automated mechanisms to increase training dataset sizes [19,20]. Another approach is to use synthetic datasets [21,22]. Hou et al. proposed a synthetic data generation and segmentation workflow which included a generative adversarial network (GAN) architecture and a nucleus segmentation CNN [21]. The workflow creates an initial set of synthetic tissue image patches and segmentation masks and then refines them using the GAN, while training a U-Net [23] segmentation model. The U-net model is trained with synthetic data with a training loss that is re-weighted over the synthetic data. This minimizes the ideal generalization loss over the real data distribution to improve the accuracy of the segmentation model. The authors demonstrated the efficacy of this approach by training a multi-cancer segmentation model and applying it to over 5000 images from 10 cancer types [24]. In another recent work, Krause et al. used conditional GAN architectures to synthesize images with and without microsatellite instability in colorectal cancer [25].

In addition to the morphological properties of individual nuclei and other micro-anatomic structures, spatial relationships between nuclei and tumor regions also play a critical role in cancer prognosis. For example, studies have shown correlations between the amount of lymphocyte infiltration in tumor regions and clinical outcome [26]. If the main objective is to quantitate such spatial interactions, patch-based methods can provide an approximation of what can be computed by pixel-level nuclei and region segmentations. Training data generation is generally less time consuming with patch-based methods; the annotator only has to label a patch as class-positive or negative (e.g., tumor-positive to indicate the patch is in or intersects a tumor region or lymphocyte-positive to indicate the patch contains lymphocytes) instead of pain-stakingly tracing the boundaries of individual nuclei and tumor regions. Le et al. employed a patch-based approach for segmentation and classification of spatial patterns of tumor regions and tumor infiltrating lymphocytes (TIL) in breast cancer WSIs [27]. Each WSI was partitioned into a grid of patches, and tumor and TIL models were trained to predict if a patch was positive. That is, in the case of tumor segmentation, the tumor model predicted if the patch was intersected a tumor region, and in the case of lymphocyte classification, the TIL model predicted if the patch contained lymphocytes. The authors showed that the predicted tumor infiltration (i.e., the ratio of lymphocyte patches intersecting tumor patches to the total number of tumor patches) correlated with survival rates.

### 3.4   Image-level classification

WSI-level or patient-level annotations are an important source of training information for image level classifications. For example, a biopsy to diagnose cancer will simply be attributed a "benign" or "malignant" label without pixel-level annotations. Such annotations are generated by Pathologists when diagnosing the disease and in the course of providing patient care; Pathologists are also typically able to quickly generate WSI-level annotations. Classification of WSIs using WSI-level annotations can be formulated as a multi-instance learning (MIL) problem to identify salient image regions. MIL methods generally partition a WSI into patches and execute an iterative two stage approach consisting of patch-level predictions carried out using a neural network followed by a data integration and image-level prediction phase.

Patch voting has been commonly used to attribute the most common predictions made at the patch-level to the WSI. Korbar et al. trained a ResNet-based architecture to predict between hyperplastic polyp, sessile serrated polyp, traditional serrated adenoma, tubular adenoma, and tubulovillous/villous adenoma on patches extracted from an H&E-stained WSI of colorectal biopsy specimens [28]. WSIs were then assigned a label according to the most common patch-level prediction, provided that the prediction was made on at least five patches. Patch-level predictions can also be aggregated by a machine learning classifier to make WSI-level inferences. Li et al. trained two ResNet-50 networks as feature extractors on patches of different sizes from breast cancer WSIs [29]. The authors utilized smaller 128x128 pixel patches to capture cell-level features and larger 512x512 pixel patches for tissue morphology and structure information. K-means clustering was performed on extracted patch feature representations to identify discriminative patches, of which histograms were generated for each WSI. A SVM then classified each WSI as normal, benign, in situ, or invasive using these histograms. Hou et al. utilized an expectation-maximization (EM) MIL approach to train models for classification of glioma and non-small cell lung carcinoma subtypes [30]. The authors assumed a hidden variable is associated with each patch that determines whether the patch is discriminative for the WSI-level label. EM was utilized to estimate this hidden variable and iteratively identify discriminative patches. Notably, to address patches close to decision boundaries that may be informative, the authors trained two CNNs at different scales in parallel and averaged their predictions.

Campanella et al. developed a patch-ranking based approach [31]. A ResNet-34 model was trained with the MIL assumption that a WSI predicted as positive has at least one patch classified as cancer. For every positive WSI, extracted patches are ranked by predicted cancer probability. As such, patches most discriminative of WSI-level positivity are iteratively learned to train the patch-based model. For WSI-level inference, the model performs predictions on extracted patches. The top $S$ most suspicious patches for cancer undergo feature extraction and are then passed onto an RNN for WSI-level prediction. One downside to ranking patches by prediction probability is that this operation is not trainable. As an alternative, the MIL problem can also be viewed from the perspective of assigning weights to patches based on their importance to the WSI label. Ilse et al. proposed a two-layer attention mechanism that provides weighted averages of input patch representations to provide trainable MIL pooling [32]. The authors utilized a breast cancer dataset, where WSIs were labeled positive if at least one breast cancer cell existed, and a colon cancer dataset, where positive WSIs contained at least one epithelial nuclei. For both tasks, the attention mechanism outperformed the traditional max and mean MIL pooling operations. Although trainable, Zhao et al. note that the weighted average approach is a linear combination operation and propose the incorporation of graph convolutional networks to generate WSI-level representations for predicting lymph node metastasis in colorectal adenocarcinoma patients[33]. Here, image patches from a WSI are each considered graph nodes. Should distances calculated between extracted features of these image patch nodes be less than a predetermined threshold, an edge is constructed. Thus, graph construction allows for generation of WSI representations that capture inter-patch relationships in a structured format.

More recently, groups have begun to incorporate RNN and Long-Short Term Memory (LSTM) networks for this final WSI inference step. To classify gastric and colonic epithelial tumors, Iizuka et al. trained the Inception-V3 model on non-neoplastic, adenoma, and adenocarcinoma image patches [34]. The trained model was used as a patch-level feature extractor. Sequences of patch features were inputted to train a RNN with two LSTM layers to predict at the WSI-level. To account for the sequence dependence of RNN-based networks, patch feature sequences were randomly shuffled during training.

### 3.5 Prediction of Patient Outcome and Treatment Response

Survival analysis is an important part of the process of predicting clinical outcomes and response to treatment for a patient or a group of patients. It aims to predict the duration of time before an event occurs such as disease remission or patient death. In the context of cancer patient survival, for example, it is used to predict the probability that a patient or a group of patients will be alive X number of months (or years) after they have been diagnosed with cancer. A variety of AI methods have been applied to survival analysis in digital pathology either using tissue images only or by integrating tissue image data and other data types.

WSISA developed by Zhu et al. is an example of methods that use tissue images only [35]. It adaptively sampled patches from the entire WSI to train CNNs. The authors argued that sampling patches from the entire image instead of annotated regions was more likely to capture and incorporate heterogeneous patterns and would provide additional information for better survival prediction. The selected patches were clustered with a K-means algorithm. Multiple CNN-based survival models were trained with patches from each cluster in order to select clusters with better prediction accuracy. The selected clusters are then aggregated via fully connected neural networks and the boosting Cox negative likelihood method. Wulczyn et al. proposed a deep learning-based analysis pipeline that transformed the survival regression problem into a classification problem. The approach divided survival time into survival intervals and predicted to which interval(s) a patient would be assigned [36]. In the analysis pipeline, each WSI was partitioned into patches randomly sampled from tissue areas. Image features were extracted from the patches through CNNs. The patch-level features were then averaged and input to a fully connected layer to classify an input image into the survival intervals. Chen et al. devised a graph-based representation of a WSI in which graph nodes are image patches and edges connect adjacent patches [37]. A graph convolutional neural network was trained with this representation of WSIs to learn hierarchical image features. The features were pooled to create WSI-level embeddings and train a model with Cox proportional loss function [38] for survival analysis. Chang et al. developed a deep learning network, called hybrid aggregation network, which aggregated data from multiple whole slide images belonging to a patient[39]. In this approach, image features are extracted at the patch-level from whole slide images using a CNN, creating whole slide feature maps. The resulting feature maps are aggregated via two aggregation components. One component was designed to aggregate features such that informative features are abstracted to region representations. The second component combined the region representations from multiple whole slide images from the same patient to predict survival rates.

Some projects combined tissue image data with other data modalities [40–44]. Mobadersany et al. proposed a method to integrate tissue image data and genomic biomarkers in a single framework to predict outcomes [42]. In this method, regions of interest were annotated in WSIs, and patches were sampled from the annotated regions. Patch-level deep features were extracted via convolutional layers. The deep features and genomic variables were then input to fully connected layers, which are connected to a Cox proportional hazards layer that models survival output. Their evaluation with lower grade glioma and glioblastoma cancer cases showed that integration of image data and genomics resulted in much better performance. Chen et al. used attention mechanisms in a multimodal deep learning framework, called Multimodal Co-attention Transformer [43]. This network learns to map image and genomic features in a latent feature space. The authors demonstrated with cases from lung, breast, brain, and bladder cancers that attention-guided integration of imaging and genomic features in a deep learning framework led to better survival prediction accuracy. Vale-Silva and Rohr developed MultiSurv, which integrates representations from clinical, imaging, and omics data modalities for cancer survival prediction [44]. MultiSurv uses submodels trained for specific data modalities to generate feature representations. The feature representations are integrated into a single representation via a data fusion layer. The integrated representation is processed by another deep learning network to output survival probabilities for a set of pre-defined time intervals. The authors used ResNext-50 CNN [45] for image data and fully connected networks for the other modalities and the integrated feature representation. The authors applied MultiSurv to 33 cancer types and showed that it produced accurate survival predictions.

## 4   Conclusions

Image analysis techniques based on traditional machine learning or statistical approaches suffered from their dependence on hand-crafted features and the need to fine-tune many method specific input parameters. While a traditional method might have performed very well on a specific dataset, it in many cases failed to easily and effectively generalize to other datasets without substantial fine tuning. For example, the organizers of the 2018 Data Science Bowl, they used traditional methods in CellProfiler[46] for a reference baseline against AI methods[47]. Even when the traditional methods were fine-tuned by experts, AI methods outperformed them by a significant margin. Similarly, evaluations done by Kumar et al.[16]

showed similar trends – CNN models surpassed the performances of traditional methods in CellProfiler and Fiji[48]. With their superior performance and better generalizability, AI methods show tremendous potential for more successful applications of tissue image analysis in research and clinical settings. As a result, the landscape of medicine is being transformed by the rapid development of new AI technologies.

## 4.1 Clinical and Research Implications

In digital pathology, deep learning methods are being used to carry out nuanced computational multiresolution tissue analyses. The goals of these analyses are to identify building blocks that can be used to create new methods for steering treatment and to act as decision support aids for improving precision and reproducibility of diagnostic classification. These methods are advancing rapidly and an ecosystem of companies is swiftly developing; in our view, over the coming decade, these methods will transform clinical and research histopathology. Nevertheless, there are challenges to be overcome in order to realize this transformation and more effectively integrate AI in clinical practice.

One of the most important challenges is the development of interpretable and explainable AI. Implementation of AI methods is still as much an art as it is engineering and science, and many AI methods are black boxes from clinicians' and biomedical researchers' perspectives. It is critical to understand how an AI model has arrived at a particular set of results in order to reliably employ AI in the decision-making process. Several research projects have started looking at explainable AI for histopathology [49,50]. Biological and environmental differences between patient cohorts can limit generalizability of conclusions reached and tools developed. Quantification of accuracy, reproducibility and generalizability of AI-based outcome predictions is an open question that needs to be further explored.

Initially, clinical decision support software that guides clinicians through recommendations in the form of predictions and prediction confidences may serve as a practical introduction of AI approaches in clinical settings. Kiani et al. developed a deep learning assistant that provides pathologists with predicted probabilities of liver cancer subtypes for hepatic tumor resections [51]. Although correct predictions significantly increased pathologist accuracy, incorrect predictions significantly decreased accuracy. As such, proper delineation of physician liability and safeguards to protect patient safety will also be of critical importance in the future integration of AI into patient care. An additional consideration centers on the capacity of AI to build upon newly acquired data. Healthcare systems are inundated with patient data on a daily basis. Regulatory mechanisms allowing for continuous learning and improvement of AI on real time patient data need to be approached with tremendous care but are an integral step in unlocking the full potential of AI in healthcare. The FDA has already been exploring the adoption of a total product lifestyle regulatory oversight to AI-based software as medical devices [52].

## 4.2 Frontiers in Algorithm and Tool Development

The complexities and types of deep learning networks are increasing – new architectures include transformers and graph networks. Some of these architectures are observed to be difficult to train, requiring large training datasets and substantial computational power. In this vein, Bommasani et al. have introduced the concept of foundation models that are trained on extremely high quantities of data then adapted towards varieties of tasks [53]. Analogous to BERT being trained on vast corpuses to learn an implicit representation of language for use in downstream tasks, one could imagine training a foundation model on a wide range and variety of images or histological specimens to be finetuned for specific applications. Feature extraction would focus more on the implicit properties of data rather than selection of those effective for the task at hand. With less dependence on task-specific loss functions, foundation model-derived representations have the potential to be finetuned for a significant range of applications – e.g.: different computer vision tasks, diseases, and organ sites. Additionally, the reduced task-specific biases of these learned representations may offer benefits in problems that require integration of temporal or multimodal data. Input specimens can be considered within reduced-bias manifolds built upon vast quantities of training data to probe relationships with and effects of data from other modalities.

In an increasing number of application areas, pre-trained models (also referred to as foundation models [53]) and domain adaptation and transfer learning methods are becoming more widely used to overcome the network complexity and computation challenges. We expect that these strategies will make their way into digital histopathology. Even now, networks pre-trained on Imagenet are widely used to speed up convergence [54]. Moreover, synthetic data generation techniques and federated training approaches could augment these strategies. With federated training, histopathology teams will have access to vast quantities of well-distributed, multi-center data even if they cannot exchange image data because of regulatory concerns as well as high cost of data exchange and storage. This shift from smaller datasets annotated for specific tasks will allow for training of more robust and accurate models. The generation of these models will invite an exciting period gearing towards practical implementation of these methodologies into clinical care with increased emphasis on logistics, regulations, risk, and most importantly, true benefit to patient health.

**Acknowledgements.**

This work was supported by the National Institutes of Health (NIH) and National Cancer Institute (NCI) grants UH3-CA225021, 3U24CA215109, and  as well as generous private support from Bob Beals and Betsy Barton.